\documentclass[letterpaper, 10 pt, conference]{ieeeconf}  

\IEEEoverridecommandlockouts                              

\usepackage{times}
\usepackage{caption}
\usepackage[pdftex]{graphicx}
\usepackage{subfigure}
\usepackage{amsmath,amssymb,amsopn,amstext,amsfonts}
\usepackage{cancel}
\usepackage[space]{cite}
\usepackage{pdfsync}
\usepackage{balance}
\usepackage{color}
\usepackage{mathtools}
\usepackage{titlesec}
\usepackage{bm}

\usepackage{diagbox}
\usepackage{float}
\usepackage{epstopdf}
\usepackage{pifont}
\usepackage{fixltx2e}
\usepackage{amsmath}
\usepackage{multirow}
\usepackage{url}
\usepackage{verbatim}
\usepackage[linkcolor=black,citecolor=black,urlcolor=black,colorlinks=true]{hyperref}
\usepackage{soul,xcolor}
\usepackage[linesnumbered,ruled,vlined]{algorithm2e}
\usepackage{overpic}
\usepackage{bbding}
\usepackage{amsmath,amssymb}
\usepackage{bbm}

\graphicspath{{./}}
\DeclareGraphicsExtensions{.png,.jpg,.eps,.pdf,.jpeg}

\vspace{-2cm}
\author{Guiyong Zheng$^{1,4,\ast}$, Jinqi Jiang$^{1,3,\ast}$, Chen Feng$^{2,\ast}$, Shaojie Shen$^{2}$, and Boyu Zhou$^{1,\dag}$
\vspace{-2cm}
\thanks{ $^{1}$School of Artificial Intelligence, Sun Yat-Sen University, Zhuhai, China.}
\thanks{$^{2}$Department of Electronic and Computer Engineering, The Hong Kong University of Science and Technology, Hong Kong, China.}
\thanks{$^{3}$School of Future Technology, Harbin Institute of Technology, Harbin, China. }
\thanks{$^{4}$School of Electronic Engineering, Xidian University, Xian, China.}
\thanks{{\tt\footnotesize $\{$cfengag, eeshaojie$\}$@ust.hk}, {\tt\footnotesize Jinqi\_J@163.com}, }
\thanks{{\tt\footnotesize zhouby23@mail.sysu.edu.cn}, {\tt\footnotesize shverses@gmail.com}}
\thanks{\textbf{$^{\ast}$ Equal Contribution $^{\dag}$ Corresponding Author}}
}
\vspace{-2cm}

\makeatletter
\def\endthebibliography{%
  \def\@noitemerr{\@latex@warning{Empty `the bibliography' environment}}%
  \endlist
}

\title{\LARGE \bf MASSTAR: A Multi-Modal and Large-Scale Scene Dataset with \\ a Versatile Toolchain for Surface Prediction and Completion}

\begin{document}

\maketitle



\begin{abstract}

Surface prediction and completion have been widely studied in various applications.
Recently, research in surface completion has evolved from small objects to complex large-scale scenes. As a result, researchers have begun increasing the volume of data and leveraging a greater variety of data modalities including rendered RGB images, descriptive texts, depth images, etc, to enhance algorithm performance. However, existing datasets suffer from a deficiency in the amounts of scene-level models along with the corresponding multi-modal information. Therefore, a method to scale the datasets and generate multi-modal information in them efficiently is essential.
To bridge this research gap, we propose MASSTAR: a \textbf{M}ulti-modal l\textbf{A}rge-scale \textbf{S}cene dataset with a ver\textbf{S}atile \textbf{T}oolchain for surf\textbf{A}ce p\textbf{R}ediction and completion. We develop a versatile and efficient toolchain for processing the raw 3D data from the environments. It screens out a set of fine-grained scene models and generates the corresponding multi-modal data. Utilizing the toolchain, we then generate an example dataset composed of over a thousand scene-level models with partial real-world data added.
We compare MASSTAR with the existing datasets, which validates its superiority: the ability to efficiently extract high-quality models from complex scenarios to expand the dataset. Additionally, several representative surface completion algorithms are benchmarked on MASSTAR, which reveals that existing algorithms can hardly deal with scene-level completion. We will release the source code of our toolchain and the dataset. For more details, please see our project page at
\href{https://sysu-star.github.io/MASSTAR}{https://sysu-star.github.io/MASSTAR}.

\end{abstract}

\section{Introduction}

\begin{figure}[t]
\centering
\vspace{0.5cm}
\setlength{\abovecaptionskip}{0.2cm}
\includegraphics[scale=0.39]{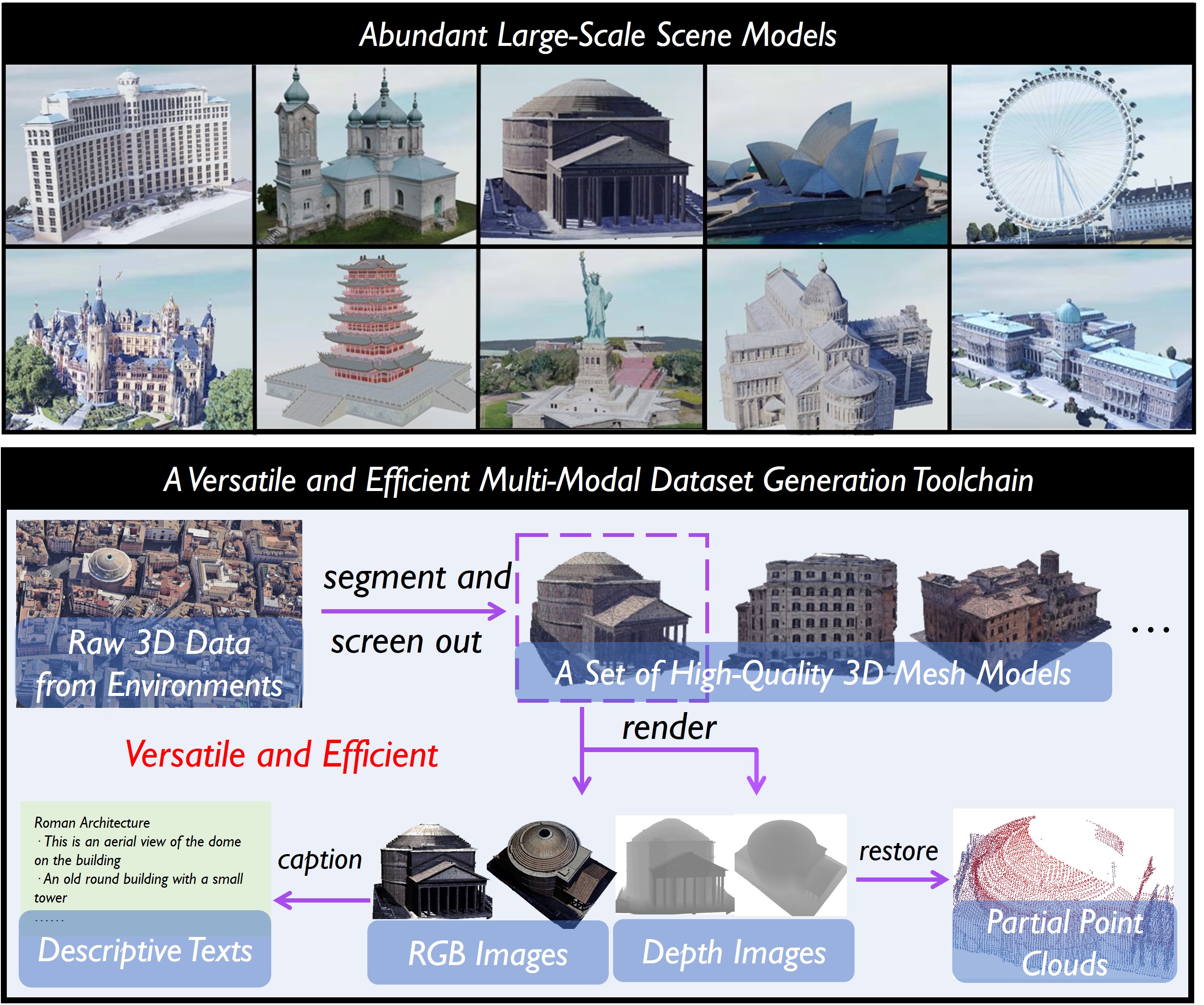}
\caption{We propose a multi-modal dataset composed of plenty of large-scale scene data for 3D surface prediction and completion, as well as a versatile and efficient toolchain to create such a dataset from raw 3D data from environments.}
\label{fig:head}
\vspace{-1.0cm}
\end{figure}

In recent years, surface prediction and completion\cite{feng2023predrecon, yuan2018pcn, aiellocross, zhang2021view, xie2020grnet, pan2021variational, chabra2020deep, chibane2020implicit} have been an essential topic in various robotic applications, including high-quality 3D reconstruction, autonomous driving and so on.
The field of artificial intelligence (AI) has witnessed significant progress, particularly with the development of generative models like Generative Adversarial Networks (GANs)\cite{goodfellow2014generative}, Diffusion Models\cite{ho2020denoising}, and Variational Autoencoders (VAEs)\cite{kingma2013auto}. These advancements have established learning-based methods as the predominant algorithms for surface completion.


Meanwhile, surface completion has also expanded its horizons toward more formidable challenges.
The main focus of existing research\cite{feng2023predrecon, yuan2018pcn, aiellocross, zhang2021view, xie2020grnet, pan2021variational,chabra2020deep,chibane2020implicit} in surface completion is not limited to small-scale objects such as chairs and tables but extends to large-scale scenes, including buildings, forests, and more. Addressing these challenges can make the algorithm more efficient and robust in the real-world application.
Recently, multi-modal learning methods and foundation model\cite{bommasani2021opportunities, radford2021learning, jun2023shap, devlin2018bert} show enhanced potential and robustness in handling complex tasks such as natural language understanding and embodied
perception of the world. This has inspired researchers to enhance surface completion\cite{radford2021learning, wu2023visual} by integrating diverse modal information such as images or texts and utilizing extensive scene-level data to train their models. 
Additionally, many surface completion algorithms\cite{yuan2018pcn, feng2023predrecon, aiellocross} are trained and tested on datasets mainly composed of synthetic models. The inherent domain gap between the real and simulated world makes such algorithms less effective when deployed in the real world.

However, existing datasets\cite{selvaraju2021buildingnet, chang2015shapenet, yao2020blendedmvs,zhang2021vipc} often fall short of addressing the challenges in this area and further hinder the research in this realm.
Primarily, many existing datasets like ShapeNet\cite{chang2015shapenet} comprise small objects like tables, sofas, etc. 
Moreover, there is a notable lack of diversity in modalities as well as the real-world data across most of these datasets\cite {selvaraju2021buildingnet,chang2015shapenet}.
Additionally, existing datasets lack a function to scale themselves efficiently and are confined to a static size.


\begin{figure}[!t]
\vspace{0.2cm}
\centering
\setlength{\abovecaptionskip}{0.2cm}
\includegraphics[scale=0.365]{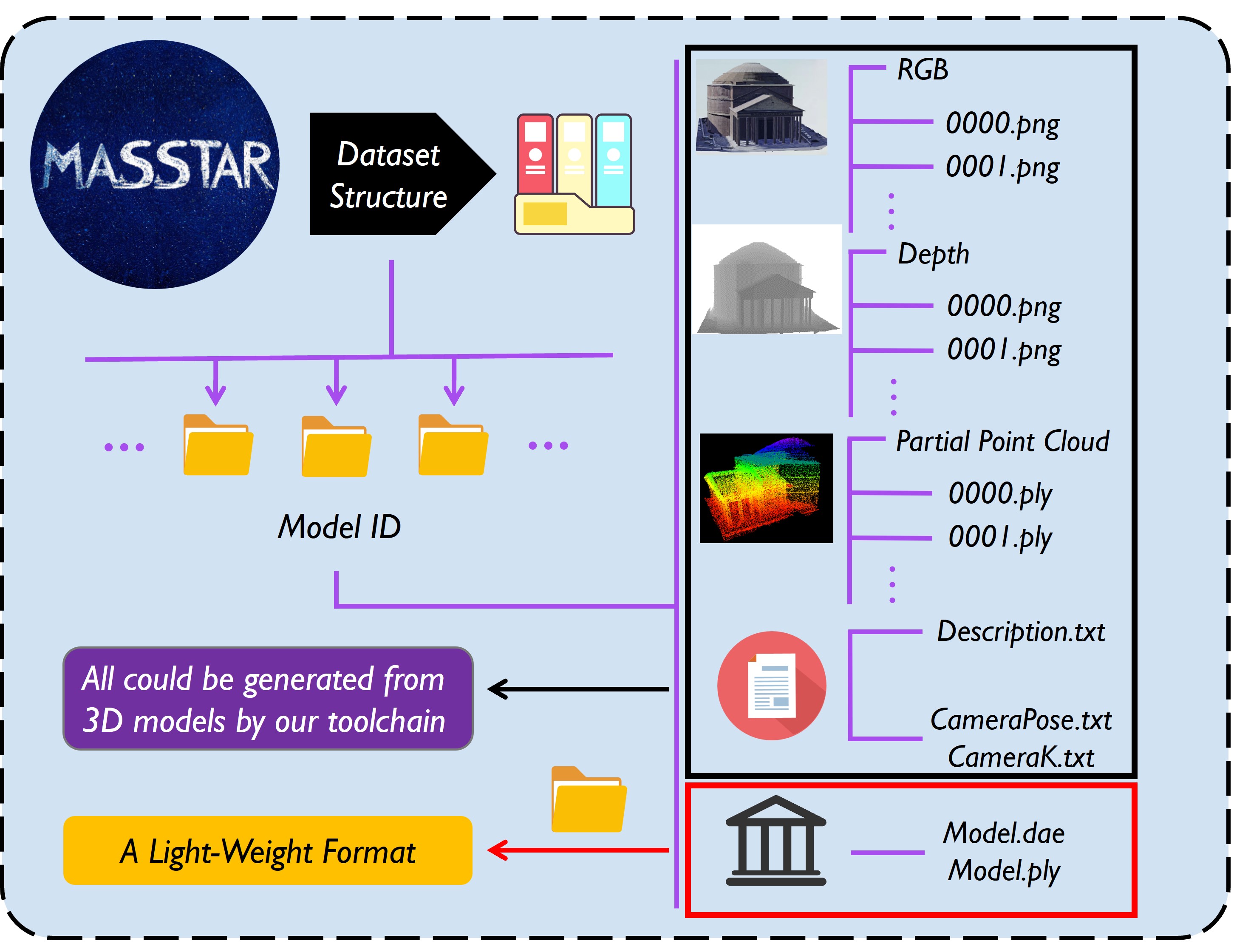}
\caption{The structure of our dataset. We release a lightweight format dataset that contains only 3D mesh models, and users can generate a complete format dataset using our toolchain.}
\label{fig:structure}
\vspace{-0.6cm}
\end{figure}

To address the issues mentioned above and to promote the development of this research field, we propose MASSTAR: a \textbf{M}ulti-modal l\textbf{A}rge-scale \textbf{S}cene dataset with a ver\textbf{S}atile \textbf{T}oolchain for surf\textbf{A}ce p\textbf{R}ediction and completion. The overall system is represented in Fig.\ref{fig:head}.
Our versatile and efficient toolchain could screen out a set of high-quality 3D mesh models from diverse raw 3D data in the environments and then produce multi-modal outputs, including images, descriptive texts, and point clouds. It operates effectively in various structured environments, such as buildings \cite{feng2023predrecon}, urban areas \cite{zhang2021continuous}, and more. Leveraging this toolchain, we create an example dataset composed of over a thousand scene-level 3D mesh models, as depicted in Fig.\ref{fig:structure}. Additionally, the toolchain could be utilized seamlessly as an independent data processing step for existing datasets like ShapeNet to facilitate surface completion tasks \cite{chang2015shapenet} as shown in Sec.\ref{sec:test}.


We compare MASSTAR with the existing and widely used datasets. Table.\ref{table:benchmark} shows the benchmark comparison of MASSTAR with the existing widely used datasets. MASSTAR outperforms because it contains scene-level models and partial real-world data, and supports four modalities. MASSTAR also provides a toolchain to generate data from environments to scale the dataset efficiently. The detailed comments on the benchmark are given in Section \ref{sec:dataset}.
We summarize our main contributions as follows:

1) We develop a versatile and efficient toolchain to screen out high-quality models from real-world or synthetic environments and generate corresponding multi-modal information.

2) We create a multi-modal large-scale scene dataset composed of over a thousand collected scene-level 3D mesh models including partial real-world data.

3) We benchmark some representative methods of surface completion and the result shows that the MASSTAR could foster research in this realm.

4) We will release the proposed toolchain and the example dataset created by the toolchain.\footnote{\url{https://github.com/SYSU-STAR/MASSTAR}}

\section{Related Work}

\subsection{\textbf{3D Data from Various Environments}}

\begin{figure}[!t]
\vspace{0.2cm}
\centering
\setlength{\abovecaptionskip}{0.2cm}
\includegraphics[scale=0.40]{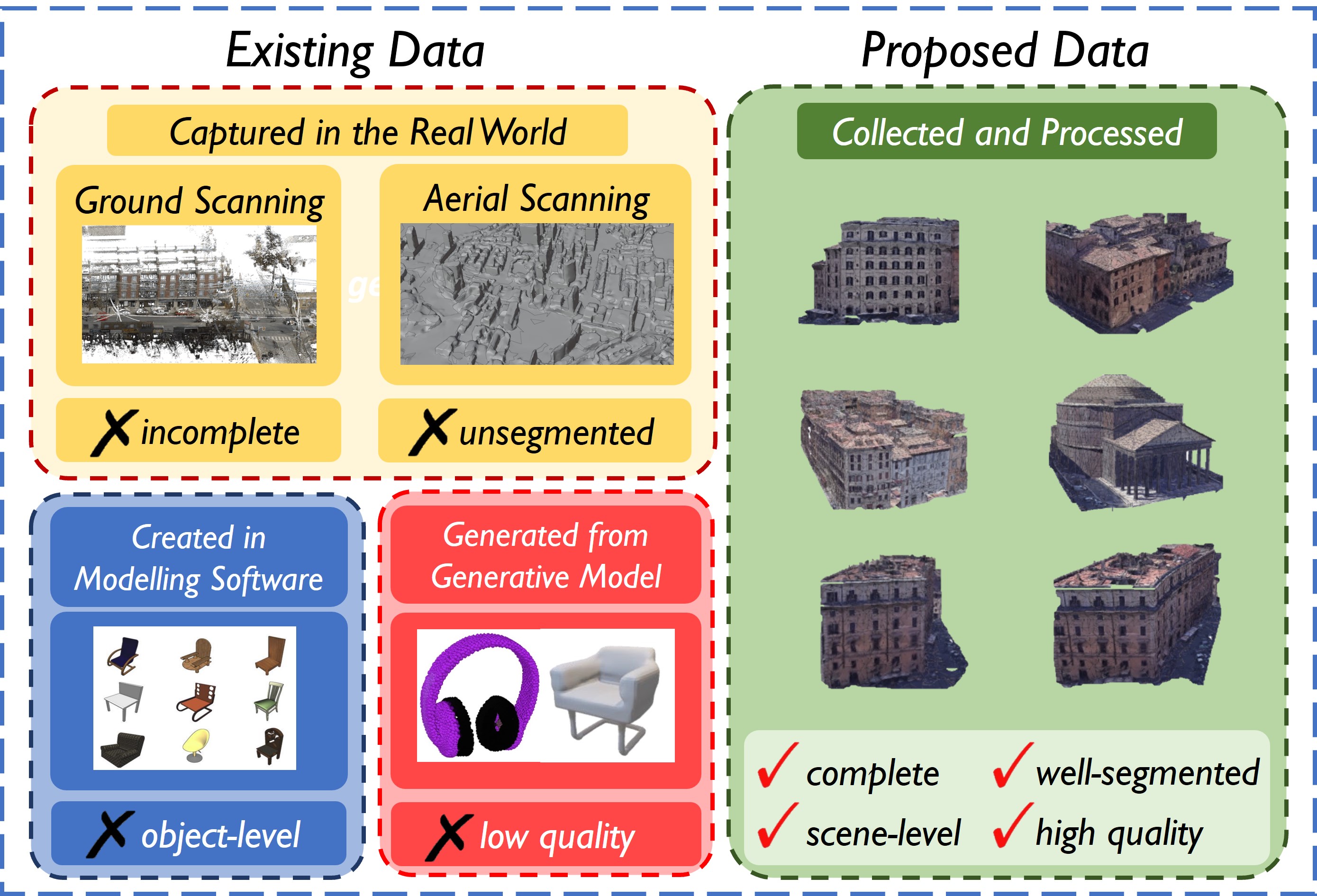}
\caption{A comparison of 3D scene models in MASSTAR with those in other datasets. While the models in the former datasets suffer from different drawbacks, the 3D scene models in MASSTAR feature complete surfaces, well-segmented scenes, object-level models, and high quality.}\label{fig:dataset}
\vspace{-0.3cm}
\end{figure}

Existing 3D data falls into two categories: data captured from the physical world and data generated within simulations. Fig.\ref{fig:dataset} shows the drawbacks of the existing datasets and the advantages of our dataset.

\vspace{0.15cm}
\noindent\textit{\textbf{1) Data Captured in the Real World}}
\vspace{0.15cm}

\begin{table}
\vspace{0.15cm}
\scriptsize
   \renewcommand\arraystretch{1.4}
   \tabcolsep=1mm
   \centering
   \caption{A comparison between MASSTAR and some existing datasets for surface completion.}
   \label{table:benchmark}
   \begin{tabular}{cccccc} 
   \hline
                     & Method  & \begin{tabular}[c]{@{}c@{}} \textbf{Ours} \end{tabular} & \begin{tabular}[c]{@{}c@{}}BuildingNet \cite{selvaraju2021buildingnet} \end{tabular} & \begin{tabular}[c]{@{}c@{}}Yao's \cite{yao2020blendedmvs} \\ \end{tabular} & \begin{tabular}[c]{@{}c@{}}ShapeNet\cite{chang2015shapenet}
                     \\\end{tabular} \\
   \hline
   \hline
   \multirow{4}{*}     & Open-source Toolchain      & \textcolor{red}{\Checkmark} & \XSolidBrush     & \XSolidBrush    & \XSolidBrush        \\ 
   \hline
                             & Modal Quantity        & \textcolor{red}{\textbf{4}} & \textbf{2} & \textbf{2}       & \textbf{3}         \\ 
   \hline
                             & Number of Models    & 1K  & 2K       & 0.1K     & \textcolor{red}{\textbf{513K}}      \\ 
   \hline
                             & Scene Level Model   & \textcolor{red}{\Checkmark}        & \textcolor{red}{\Checkmark}          & \textbf{partial}      & \XSolidBrush     \\ 
   \hline

                            & Real-world Data  & \textcolor{red}{\textbf{partial}}        &\XSolidBrush           & \XSolidBrush      & \XSolidBrush     \\ 
   \hline
   \end{tabular}
   \vspace{-1.5cm}
\end{table}

Benefiting from the development of sensors and reconstruction algorithms\cite{xu2021fast,xu2022fast,lindenberger2021pixel}, the 3D data captured from the real world could maintain higher fidelity with a much cheaper device. We categorize real-world data into two distinct groups: ground-scanned data and aerial-captured data, owing to the distinct methods of data collection.

\noindent\textbf{The 3D data from ground scanning.} It is primarily acquired from lidar sensors or RGB-D cameras operated by ground-based robots, workers, or researchers \cite{lin2022r, geiger2013vision, sturm2012benchmark}. This type of data typically boasts high resolution, fidelity, and detailed geometry. However, since the devices are constrained on the ground, capturing the full view of the target scene is difficult, often resulting in inadequate scanning of structures such as roofs. The incomplete scene poses a notable challenge for surface completion tasks, as it fails to furnish a comprehensive scene model that could serve as the definitive benchmark for supervising algorithm training.

\noindent\textbf{The 3D data from aerial scanning.} It's a significant subset of our dataset, and primarily originates from high-definition cameras mounted on either manned or unmanned aerial vehicles\cite{AgEagle-web,esri-web, PIX4Dcloud-web}. These datasets are expansive and provide abundant, texture-rich information. However, it's important to note that researchers cannot readily employ these datasets in their raw form due to the absence of segmentation. The unsegmented nature of the aerially captured 3D data poses a substantial challenge. Segmentation is a crucial preprocessing step in harnessing the potential of these datasets for various applications. Currently, researchers often need to employ sophisticated algorithms and techniques to partition this data into meaningful segments, such as objects, buildings, or regions of interest. This segmentation process not only aids in organizing the data but also lays the foundation for subsequent processes and model training tasks.

\vspace{0.15cm}
\noindent\textit{\textbf{2) Data Created in Modeling Software}}
\vspace{0.15cm}

The 3D data included in these datasets \cite{selvaraju2021buildingnet, chang2015shapenet} have been meticulously crafted utilizing advanced 3D modeling software, including CAD\footnote[1]{\url{https://www.autodesk.com.cn}}, Blender\footnote{\url{https://www.blender.org}}, and Unreal Engine\footnote{\url{https://www.unrealengine.com}}. However, it's important to note that generating expansive 3D scenes with a high level of detail can be a formidable task. As a result, researchers typically concentrate their efforts on creating object-level 3D models, such as chairs, tables, sofas, and a variety of other items.

\vspace{0.15cm}
\noindent\textit{\textbf{3) Data from Generative Models}}
\vspace{0.15cm}

In addition to manually creating data, the emergence of generative models has expanded the scope of dataset creation. This innovation has unlocked the potential for novel approaches to 3D data generation. Pioneering 3D generation methods \cite {xu2023dream3d,nichol2022point,cheng2023sdfusion} have been developed to create 3D data from diverse sources, including single-view images and textual descriptions, among others. However, despite the promise of AI-driven 3D generation, the development and training of generative models heavily rely on the availability of high-quality training data. Consequently, the limited supply of suitable datasets continues to hinder the full realization of 3D data generation using generative models.

\subsection{\textbf{Toolchain for Multi-Modal Data Generation}}

It is very common to process datasets with different tools to obtain data in other modalities. For example, employing 3D rendering software to obtain images from 3D mesh models is a widespread practice in the fields of multi-view stereo (MVS)\cite{yao2020blendedmvs}, structure from motion (SFM)\cite{choy20163d}, and multi-modal neural network models\cite{nichol2022point}. Besides, the remarkable advancement of large-scale AI models in recent years has promoted a growing number of researchers to leverage AI for automated data annotation, including model segmentation, image segmentation, and text annotation, among others.

However, existing methods are often limited to a single application or a specific scenario, and users are required to utilize them separately to accomplish various tasks, which can be a cumbersome and time-consuming process. Presently, there is a noticeable absence of versatile tools capable of seamlessly automating the generation of multi-modal data. The emergence of large-scale AI models makes a versatile and efficient toolchain that could be used to process data in various scenarios due to their amazing zero-shot performance.

\begin{figure}[!t]
\vspace{0.2cm}
    \centering
    \setlength{\abovecaptionskip}{0.2cm}
    \captionsetup{type=figure}
    \includegraphics[scale=0.40]{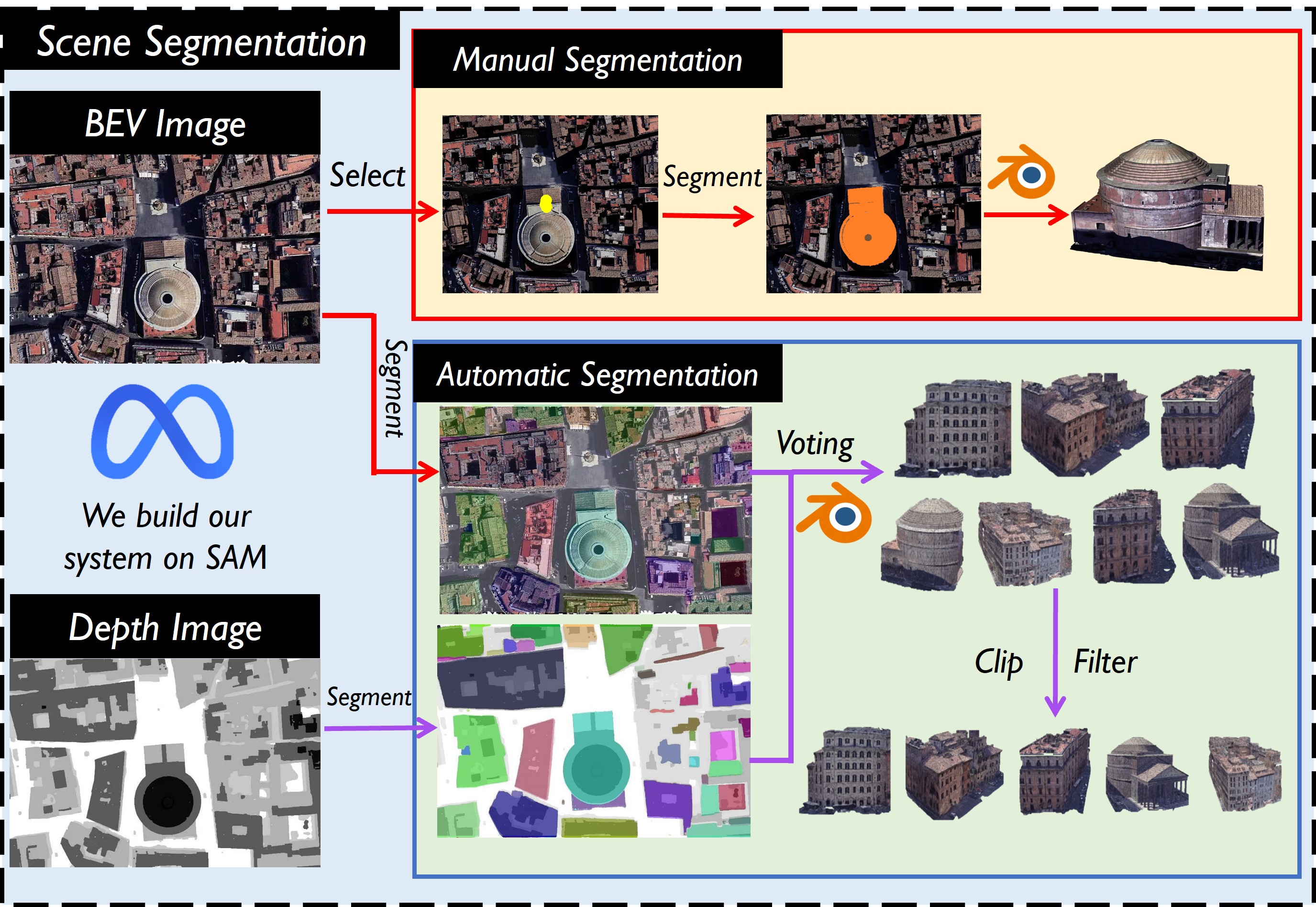}
    \caption{An overview of 3D scene segmentation. Initially, we generate the depth image and RGB image by rendering a bird's-eye view of each scene. Users have the option to employ SAM\cite{kirillov2023segany} for segmenting top-view images in manual mode or automatic mode. Subsequently, the 3D mesh model is sliced using Blender, and then CLIP\cite{radford2021learning} is utilized to filter out non-architectural categories.}\label{fig:segmentation}
    \vspace{-0.7cm}
\end{figure}

\section{Multi-Modal Dataset Generation Toolchain}
\label{sec:tc}

In this section, we will introduce our toolchain, divided into four parts: 3D scene segmentation (Sect.\ref{sec:sam}), images rendering (Sect.\ref{sec:render}), description texts generation  (Sect.\ref{sec:caption}), and partial point clouds generation (Sect.\ref{sec:partial}). The structure of the entire toolchain is depicted in Fig.\ref{fig:head}.

\subsection{\textbf{3D Scene Segmentation}}
\label{sec:sam}
The 3D scene segmentation part is to screen out a set of high-quality 3D mesh models from various environments as shown in Fig.\ref{fig:segmentation}. Users could choose to split the 3D scene manually or automatically.

We adopt an advanced AI model called Segment Anything Model (SAM) \cite{kirillov2023segany}, which has shown excellent performance in segmenting complex and diverse images with the ability to perform panoramic segmentation. In the first stage, we utilize it to segment the bird's eye view image of the 3D scene and get the mask of the buildings on the image. Then in the second stage, the 3D scene is segmented by Blender according to the mask returned by the first stage.

In automatic segmentation mode, we first use SAM to segment the bird's-eye view images of the 3D scene and generate some segmented models using Blender. Since SAM has no semantic information about the scene, we encounter some non-building models (such as roads, trees, etc.). To address this issue, we leverage the Contrastive Language-Image Pre-training (CLIP) model, which is a powerful multi-modal model that aligns images with text and performs well in zero-shot tasks. Specifically, CLIP encodes images and texts into separate vectors and evaluates their similarity by computing the distance between the two vectors. We utilize CLIP to evaluate the relevance of certain segmented models to buildings, effectively removing any unrelated structures

For manual segmentation mode, users are required to provide a point or rectangle prompt information for the region of interest on the bird's-eye view image. SAM utilizes this prompt to generate the mask of the buildings on the image, and then the 3D scene is segmented by Blender according to this mask.

\begin{figure}[!t]
\vspace{0.2cm}
    \centering
    \setlength{\abovecaptionskip}{0.2cm}
    \captionsetup{type=figure}
    \includegraphics[scale=0.39]{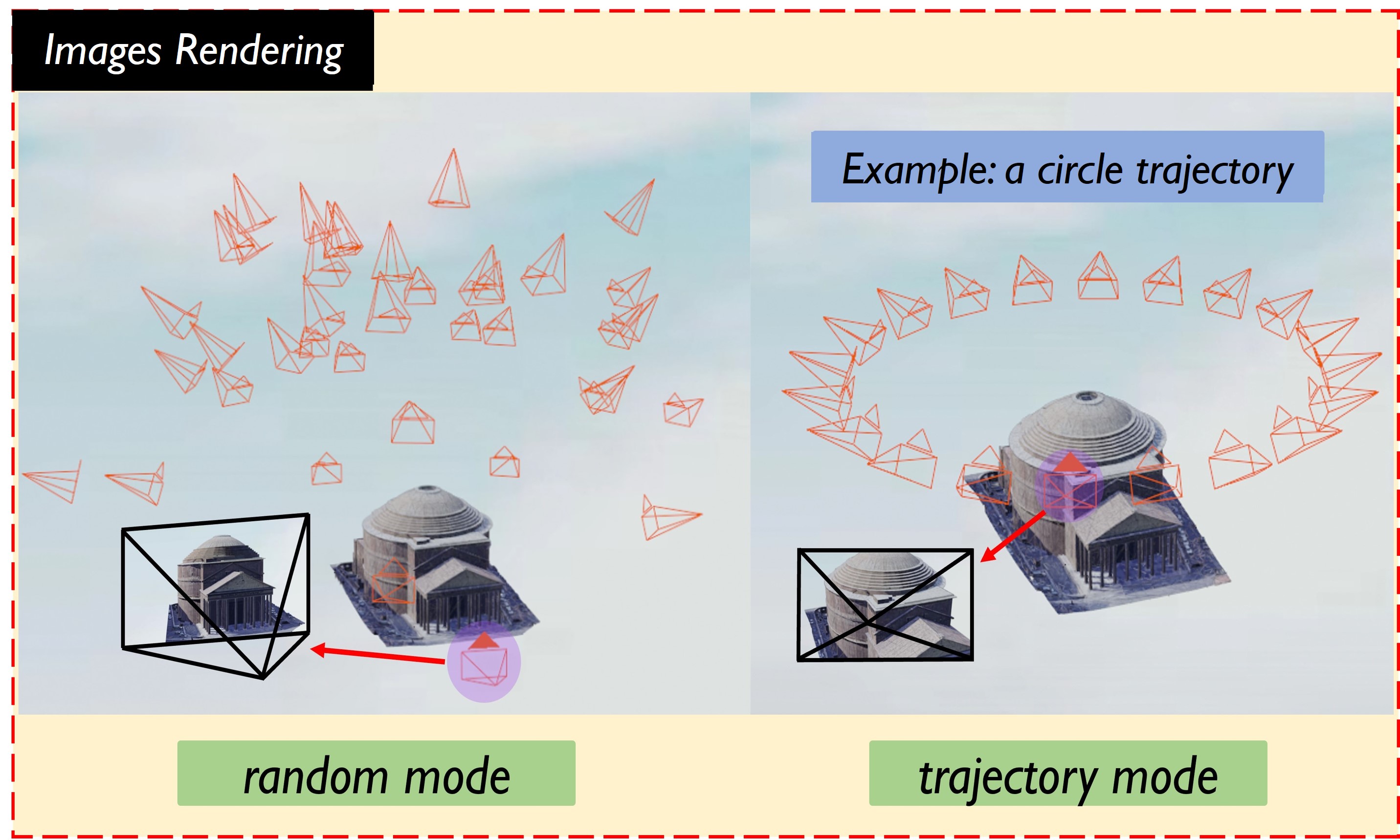}
    \caption{An example of the image rendering part of the toolchain. We offer the random mode (left) and trajectory mode (right) for users.}\label{fig:render}
    \vspace{-0.8cm}
\end{figure}

\subsection{\textbf{Images Rendering}}
\label{sec:render}

Our image rendering component utilizes Blender, an open-source software celebrated for its 3D format support and efficient rendering engine.
This component renders 3D scenes efficiently from user-defined viewpoints which encompass two distinct modes: random and trajectory. This allows users to generate RGB or depth images according to their specific requirements.

Our framework begins by normalizing each 3D scene within a bounding cube. It then creates user-defined lighting configurations and utilizes Blender's advanced rendering engine to generate the desired images at last.
The random mode ensures a diverse selection of viewpoints by completely randomizing camera angles, thereby helping to prevent potential overfitting during the training process. In contrast, the trajectory mode allows users to input a 5-dimensional trajectory, covering parameters $[x, y, z, pitch, yaw]$. Fig.\ref{fig:render} illustrates a rendering example demonstrating the trajectory mode, using a circular path as a sample input.

\subsection{\textbf{Descriptive Texts Generation}}
\label{sec:caption}

Recent advancements in Large Vision-Language Models (LVLM) have enabled seamless extraction of textual descriptions directly from images themselves \cite{li2022blip, zhang2023llamaadapter, gao2023llamaadapterv2}. BLIP \cite{li2022blip} represents one such LVLM framework, proficient in both vision-language understanding and generation tasks. Pre-trained on a vast dataset of 14 million samples, BLIP exhibits remarkable zero-shot capabilities, making it well-suited for generating building description text.

\begin{figure}[!t]
\vspace{0.2cm}
    \centering
    \setlength{\abovecaptionskip}{0.2cm}
    \captionsetup{type=figure}
    \includegraphics[scale=0.355]{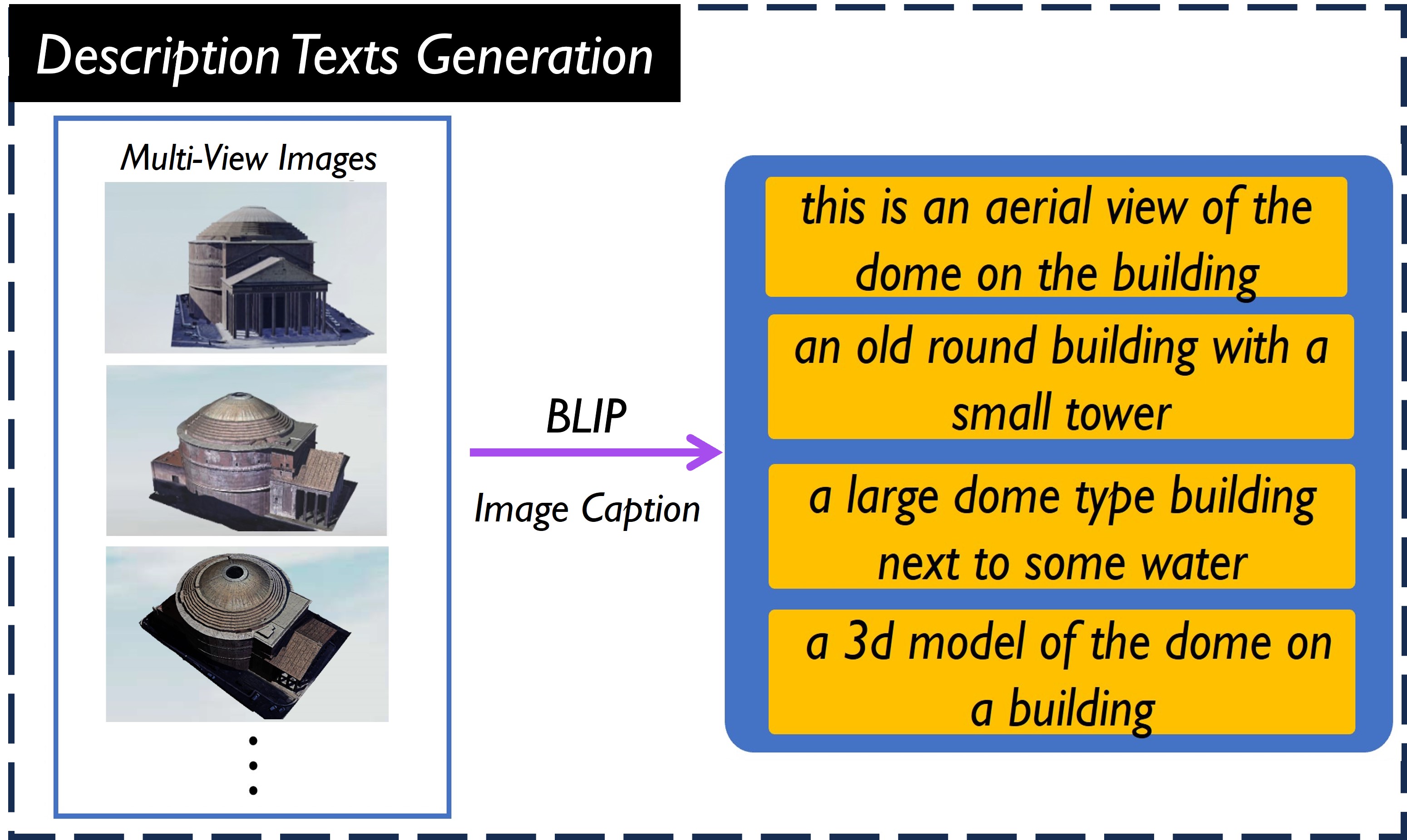}
    \caption{An example of the descriptive texts rendering part of the toolchain. BLIP is employed \cite{li2022blip} to perform zero-shot image-to-text generation.}\label{fig:text}
    \vspace{-1.0cm}
\end{figure}

In our toolchain, we leverage BLIP \cite{li2022blip} for zero-shot image-to-text generation. To ensure the quality and coherence of the text descriptions, we integrate images rendered from multiple perspectives during the text generation process. Specifically, we employ the pre-trained checkpoint of BLIP \cite{li2022blip} with ViT-B, a variant tailored for Image-Text Captioning tasks, using default parameters. Fig.\ref{fig:text} illustrates an example of this process.

\subsection{\textbf{Partial Point Clouds Generation}}
\label{sec:partial}

To enhance the effectiveness of downstream tasks related to point cloud completion and prediction, we furnish standard partial point cloud data. Due to the rendered depth images covering only a small portion of the model, we randomly combine 1-3 depth images from different viewpoints to back-project and generate more comprehensive partial point clouds. Specifically, we utilize the intrinsic parameters of the camera and the extrinsic parameters of the camera in the world coordinate system to perform back-projection from the multi-view depth images, producing partial point clouds.


\begin{figure*}[h]
\vspace{0.2cm}
    \centering
    \setlength{\abovecaptionskip}{0.2cm}
    \captionsetup{type=figure}
    \includegraphics[scale=0.42]{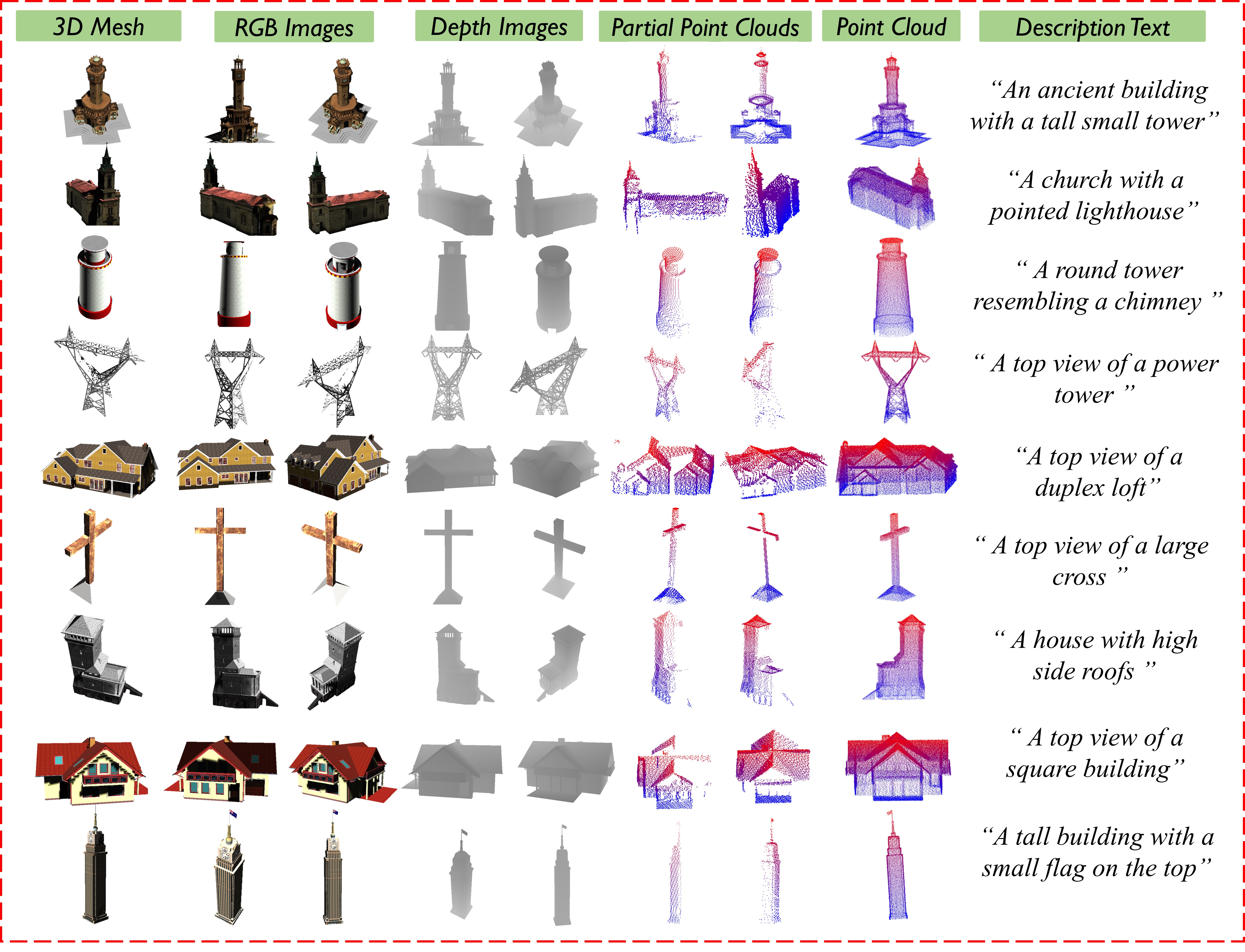}
    \caption{ Some examples of independent data processing using the proposed toolchain on some existing datasets\cite{selvaraju2021buildingnet,chang2015shapenet}.}\label{fig:example}
    \vspace{-0.5cm}
\end{figure*}

\section{Large-Scale 3D Scene Dataset}
\label{sec:dataset}

We assemble our multi-modal surface completion dataset by collecting raw 3D data from a wide range of open-source projects and resources, primarily incorporating models from UrbanScene3D\cite{liu2021urbanscene3d} and Sketchfab\footnote[1]{https://sketchfab.com/} under the CC license. UrbanScene3D offers several urban-level 3D data created in Airsim, which are ideal raw data for surface completion tasks. Additionally, 3D urban-level data from Sketchfab that is captured in the real world is either collected. Then, we utilize our toolchain to process all the raw data and generate a multi-modal dataset. It is important to highlight that during the 3D raw data segmentation process, only 15\% of the data is segmented manually.

Each model in the dataset is well segmented with fine geometric and abundant texture. Table.\ref{table:benchmark} shows a comparison of our MASSTAR and other datasets. MASSTAR achieves the first dataset composed of multi-modal scene-level models with an open-source and versatile toolchain for users to build such a dataset from any raw 3D data or traditionally insufficiently augmented dataset. Moreover, MASSTAR stands out by integrating more real-world data models, offering richer detail and complexity, presenting a tougher challenge for the surface completion algorithm.


\section{Experiment}
In this section, we conduct two experiments including processing the existing insufficient augmented datasets using the toolchain and a benchmark between some representative surface completion algorithms.

\subsection{\textbf{Versatility of the Toolchain}}
\label{sec:test}

To showcase the versatility of our proposed toolchain, we conduct plugin-in experiments on some samples of existing datasets including BuildingNet\cite{selvaraju2021buildingnet} and ShapeNet\cite{chang2015shapenet} to generate multi-modal information. The result is illustrated in Fig.\ref{fig:example}. Since these models are intricately created in synthetic environments, there is no need to leverage the segmentation part to screen out the models from raw 3D data. Through this experiment, we find that the proposed toolchain could be used as an independent data processing step on existing datasets seamlessly. It highlights the versatile capability of our toolchain.
\vspace{-0.3cm}
\subsection{\textbf{Surface Completion Benchmark}}
\label{sec:exp}

We conduct a comparative analysis of three surface prediction and completion algorithms, including SPM\cite{feng2023predrecon}, PCN\cite{yuan2018pcn}, and XMFnet\cite{aiellocross}. PCN is an encoder-decoder framework that completes a partial input point cloud with a typical coarse-to-fine scheme. It is the first point cloud prediction model that does not require any shape assumptions and most point cloud prediction models are based on PCN. SPM is specifically designed and developed for robotic systems based on PCN. It significantly improves prediction accuracy and inference speed by retaining the original input point cloud and employing a simplified network design.  In contrast, XMFnet attempts to incorporate image information to enhance prediction. It adopts a multi-modal point cloud completion strategy, taking partial point cloud and image as input, leveraging a transformer framework to directly fuse at a feature level for improved prediction efficacy.

\vspace{0.15cm}
\noindent\textit{\textbf{1) Implementation Details}}

Our MASSTAR contains 1027 models. For each model, we generated 15 partial point clouds, 15 RGB images, and 15 corresponding depth images from different viewpoints, along with a complete point cloud as ground truth.
We set 2048 input points for each partial point cloud and 2048 points for each ground truth, with the image resolution set to 224 $\times$ 224 pixels, consistent with the settings used in XMFnet\cite{aiellocross}. We then randomly partition the dataset into two parts, with 70\% allocated as training data and 30\% allocated as testing data. As for more details, we train all the models for 400 epochs and test them on a single NVIDIA RTX 4090.

\begin{table}
\vspace{0.15cm}
   \renewcommand\arraystretch{1.4}
   \tabcolsep=1mm
   \centering
   \caption{Time efficiency and resource consumption of each surface completion method tested on MASSTAR.}
   \label{table:efficiency}
   \begin{tabular}{l|ccc} 
   \hline
                   Evaluation Metric  & SPM\cite{feng2023predrecon}  & PCN\cite{yuan2018pcn} & XMFnet\cite{aiellocross}  \\
   \hline
   \hline
   \multirow{1}{*}{{Inference Time (ms)}}     & 1.61       &\textbf{0.506}         & 210              \\ 
   \hline
   \multirow{1}{*}{{FLOPs (G)}}     &4.06       &\textbf{3.32}          &387.68              \\ 
   \hline
    \multirow{1}{*}{{Param (MB)}}     &9.73       &\textbf{4.11}           &8.73             \\ 
   \hline
   \end{tabular}
   \vspace{-0.8cm}
\end{table}

\vspace{0.15cm}
\noindent\textit{\textbf{2) Evaluation Metrics}}


We introduce a comprehensive evaluation metric for surface prediction and completion algorithms, which focuses on two main aspects: resource utilization and prediction quality.

\noindent\textbf{Resource Utilization.} The efficiency of resource utilization entails both inference time and memory footprint. They are critical factors for onboard computing robots with limited computational resources in real-world applications.

\noindent\textbf{Prediction Quality.} To assess prediction quality, we expand upon the method outlined in \cite{knapitsch2017tanks}. This method computes several indices, including Chamfer Distance (CD), precision, recall, F-score, and Area Under the Curve (AUC).

Here we denote the reconstructed point cloud as $\mathcal{P}$ and the ground truth as $\mathcal{Q}$. We could formulate the distance between a point $p$ in $\mathcal{P}$ and $\mathcal{Q}$ concerning \cite{knapitsch2017tanks} as:
	 \begin{equation}
	     {\rm dist}(p, \mathcal{Q}) =\mathop{{\rm min}}\limits_{q \in \mathcal{Q}}\Vert p-q \Vert. 
	 \end{equation}
Subsequently, we calculate the CD using two different distance definitions. Specifically, when the distance $dist$ is computed using the L1 norm, it is labeled as L1-CD. Similarly, if the L2 norm is utilized, it is labeled as L2-CD.
	 \begin{equation}
      \begin{aligned}
	     {\rm CD}&=\frac{1}{2|\mathcal{P}|}\sum\limits_{p \in \mathcal{P}}{{\rm dist}(p,\mathcal{Q})}+\frac{1}{2|\mathcal{Q}|}\sum\limits_{q \in \mathcal{Q}}{{\rm dist}(q, \mathcal{P})}.
      \end{aligned}
	 \end{equation}
Moreover, precision $\rm P(\tau)$ and recall $\rm R(\tau)$ can be employed to evaluate the results, where the parameter $\tau$ represents the predefined distance threshold.
	 \begin{equation}
	     \begin{aligned}
           {\rm P}(\tau) &= \frac{1}{|\mathcal{P}|}\sum\limits_{p \in \mathcal{P}}{\mathbbm{1}({\rm dist}(p,\mathcal{Q}) < \tau)} \\
           {\rm R}(\tau) &= \frac{1}{|\mathcal{Q}|}\sum\limits_{q \in \mathcal{Q}}{\mathbbm{1}({\rm dist}(q,\mathcal{P}) < \tau)},
        \end{aligned}
    \end{equation}

where $\mathbbm{1}(\cdot)$ represents the indicator function.

The F-score ${\rm F}(\tau)$ is the harmonic mean of precision $\rm P(\tau)$ and recall $\rm R(\tau)$ at the threshold $\tau$, as follows:

	 \begin{equation}
	     \begin{aligned}
	      {\rm F}(\tau) &= \frac{2{\rm P}(\tau){\rm R}(\tau)}{{\rm P}(\tau) + {\rm R}(\tau)}.
	     \end{aligned}
	 \end{equation}
  
Finally, we introduce F-score curves at various thresholds $\tau$ to compute the AUC and comprehensively evaluate the prediction quality.  
	 \begin{equation}
	     \begin{aligned}
	     {\rm AUC}=\int_{{\tau}_{1}}^{{\tau}_{2}} {\rm F}(\tau) d\tau
	     \end{aligned}
	 \end{equation}
  
To enhance clarity, we utilize a logarithmic scale for the x-axis, with the AUC calculated correspondingly.

\begin{table}[t]
\vspace{0.3cm}
   \renewcommand\arraystretch{1.4}
   \tabcolsep=1mm
   \vspace{0cm}
   \centering
   \setlength{\abovecaptionskip}{0.2cm}
   \caption{The quality of prediction results of each surface completion method tested on MASSTAR.\label{table:result}}
   \begin{tabular}{l|ccc} 
   \hline
                   Evaluation Metric  & SPM\cite{feng2023predrecon}  & PCN\cite{yuan2018pcn} & XMFnet\cite{aiellocross}  \\
   \hline
   \hline
   \multirow{1}{*}{L1-CD $(\times10^{-3})$ $\downarrow$}     &31.34       &32.70         &$\textbf{29.75}$              \\ 
   \hline
   \multirow{1}{*}{L2-CD $(\times10^{-3})$ $\downarrow$}     &4.97      &5.33          &$\textbf{2.67}$               \\ 
   \hline
    \multirow{1}{*}{Precision $\uparrow$}     &$\textbf{0.772}$       &0.613           &0.735               \\ 
   \hline
    \multirow{1}{*}{Recall $\uparrow$}     &0.566       &0.571         &$\textbf{0.593}$              \\ 
   \hline
    \multirow{1}{*}{F-score $\uparrow$}     &0.642        &0.589          &$\textbf{0.655}$              \\ 
   \hline
    \multirow{1}{*}{AUC $\uparrow$}     &0.604     &0.554           &$\textbf{0.614}$              \\ 
   \hline
   \end{tabular}
   \vspace{-1.6cm}
\end{table}

\vspace{0.2cm}
\noindent\textit{\textbf{3) Results and Discussion}}

\begin{figure*}[!t]
    \vspace{0.2cm}
    \centering
    \setlength{\abovecaptionskip}{0.2cm}
    \captionsetup{type=figure}
    \includegraphics[scale=0.460]{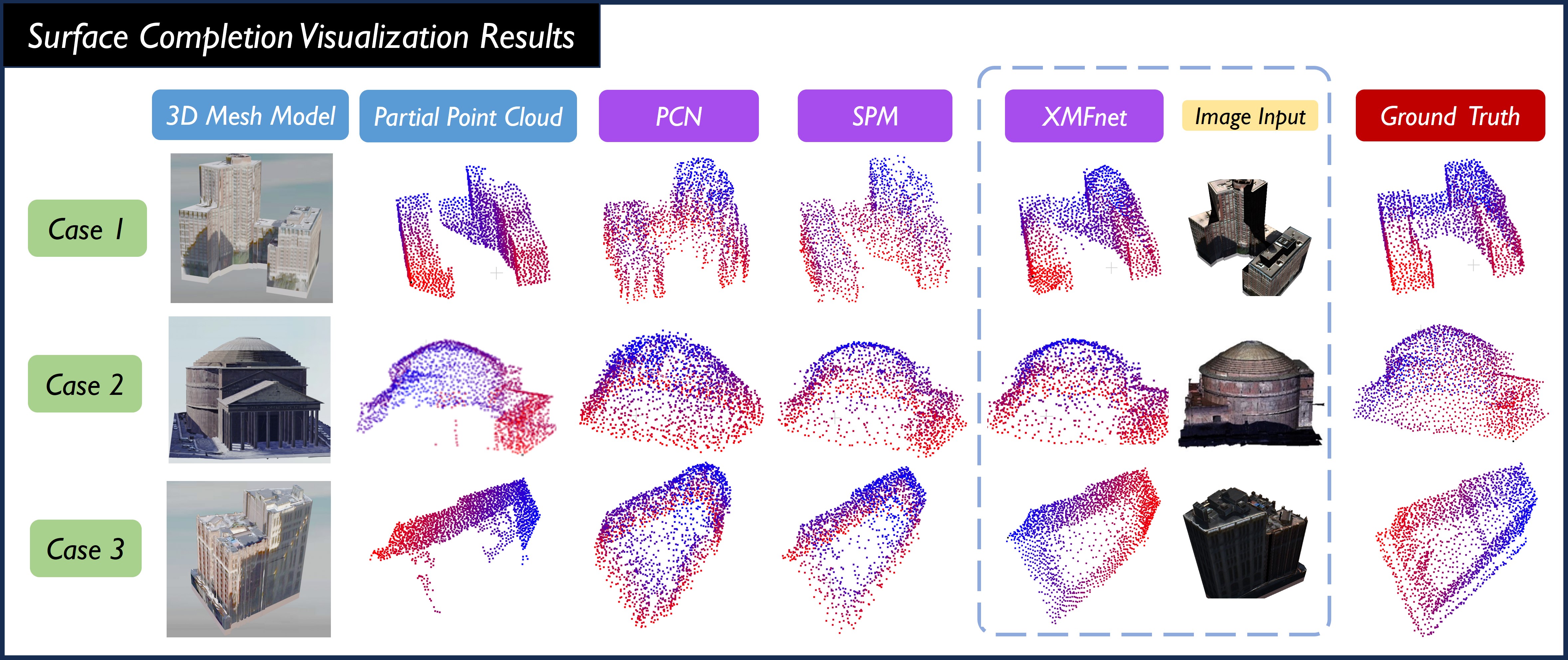}
    \caption{ A visualization comparison of the three surface prediction and completion methods in three sampled cases.}\label{fig:result}
    \vspace{-0.25cm}
\end{figure*}

\noindent\textbf{Time Efficiency and Resource Consumption.} We compared the time efficiency and resource consumption of different algorithms with indices including inference time, storage consumption, and model parameters.

As shown in Table.\ref{table:efficiency}, which benefits from a lightweight network structure, the PCN and SPM take less inference time than XMFnet. This superiority makes it promising to employ PCN and SPM on onboard-computed robots.

\noindent\textbf{Prediction Results on MASSTAR.} The prediction results are of vital importance for surface prediction and completion algorithms. We compare several indices, including L1-CD, L2-CD, F-score, precision, recall, and AUC. 
The reconstruction results are presented in Table.\ref{table:result}, with thresholds $\tau$ set to 0.001 and the integration range in AUC calculation set to [0.0001, 0.01].  Fig.\ref{fig:result} presents a visualization of the results.

Fig.\ref{fig:result} demonstrates that XMFnet exhibits superior visualization quality. This can be attributed to the retention of the local point cloud from the input, resulting in relatively high accuracy for this method. Additionally, XMFnet places greater emphasis on overall shape considerations, leading to better recall performance. Leveraging a Transformer structure, XMFnet effectively integrates images and partial point clouds, resulting in the lowest L1-CD and L2-CD metrics.

The SPM has the highest precision mainly because it will remain the exact input point cloud in the output. Moreover, PCN struggles to capture geometric details within such a large-scale scene and results in the lowest precision due to its simplistic network structure.

\begin{table}[h]
\vspace{-0.1cm}
   \renewcommand\arraystretch{1.4}
   \tabcolsep=1mm
   \vspace{0cm}
   \centering
   \setlength{\abovecaptionskip}{0.2cm}
   \caption{The quality of prediction results of each surface completion method tested on ShapeNet-VIPC.\label{table:result_shapenet}}
   \begin{tabular}{l|ccc} 
   \hline
                   Evaluation Metric  & SPM\cite{feng2023predrecon}  & PCN\cite{yuan2018pcn} & XMFnet\cite{aiellocross}  \\
   \hline
   \hline
   \multirow{1}{*}{L1-CD $(\times10^{-3})$ $\downarrow$}     &23.72      &24.77         &$\textbf{22.69}$               \\ 
   \hline
    \multirow{1}{*}{L2-CD $(\times10^{-3})$ $\downarrow$}     &1.62      &1.79         &$\textbf{1.41}$              \\ 
   \hline
    \multirow{1}{*}{Precision $\uparrow$}     &0.864       &0.839           &$\textbf{0.867}$              \\ 
   \hline
    \multirow{1}{*}{Recall $\uparrow$}     &0.736       &0.729         &$\textbf{0.761}$              \\ 
   \hline
    \multirow{1}{*}{F-score $\uparrow$}     &0.793        &0.778          &$\textbf{0.810}$               \\ 
   \hline
    \multirow{1}{*}{AUC $\uparrow$}     &0.721     &0.702           &$\textbf{0.741}$              \\ 
   \hline
   \end{tabular}
   \vspace{-0.5cm}
\end{table}

\noindent\textbf{Prediction Results on ShapeNet-ViPC\cite{zhang2021vipc}.} Table.\ref{table:result_shapenet} illustrates the performance of these methods tested on ShapeNet-ViPC, which is a dataset consisting of multi-view images and point clouds rendered from a subset of models in ShapeNet. This dataset has been previously utilized by XMFnet and others\cite{zhang2021vipc, aiellocross}. We train and test on ShapeNet-ViPC, using the same training setup as on MASSTAR. Through comparing the corresponding figures in Table.\ref{table:result} and Table.\ref{table:result_shapenet}, we could find that the performance observed with MASSTAR is inferior to that seen on ShapeNet-ViPC. It indicates the heightened complexity and challenges presented by the scene-level models in MASSTAR. Moreover, a notable disparity in algorithmic performance on MASSTAR, exceeding that on ShapeNet-ViPC, is observed. This disparity underscores the efficacy of the scene-level models within the MASSTAR dataset in delineating the performance distinctions across various algorithms.

\noindent\textbf{Discussion on Future Research of Surface Completion.}
Through a comprehensive benchmark analysis and investigation of related work, it turns out that existing surface completion techniques can be divided into two categories.
The first category focuses on quality. These methods\cite{aiellocross,jun2023shap} prioritize accuracy over speed or resource efficiency and are mainly designed for use on personal computers.
The second category is efficiency-centered. These techniques\cite{yuan2018pcn,feng2023predrecon} have been designed to support robotics applications, with their efficiency being a critical determinant.
Thanks to the availability of various sensors like cameras and lidar, robots could gather multi-modal data affordably. As a result, multi-modal approaches are anticipated to play a significant role in surface completion tasks within robotics.
In future research, the primary focus will be exploring ways to incorporate multi-modal learning into robotics that can enhance the performance of surface completion without significantly compromising efficiency.

\vspace{0.15cm}
\section{Conclusion and future work}
\label{sec:caf}

\vspace{0.15cm}
\subsection{\textbf{Conclusion}}

This paper introduces a versatile toolchain capable of producing multi-modal data from synthetic or real-world environments. Through this toolchain, a large-scale, multi-modal scene dataset is established for surface completion tasks. We also test our toolchain in some samples of existing datasets including ShapeNet and BuildingNet. Finally, three surface completion methods are trained and tested using the proposed dataset and compare the results with those on ShapeNet-ViPC. A comprehensive discussion about the results as well as future research on surface completion is provided.

\vspace{0.05cm}
\subsection{\textbf{Limitations and Future Work}}

There are still some limitations to MASSTAR that could be improved in the future.

1) There is only an insufficient number of 3D scene-level models in MASSTAR. We plan to build a website that allows users to create a crowdsourced dataset by uploading the scene-level models they collect to the website.

2) The modalities in MASSTAR are still limited, mainly including 3D mesh models, images, descriptive texts, and partial point clouds. We plan to develop more modalities such as sketches or semantic point clouds in the future.

3) With the development of AI technology, we anticipate the emergence of more powerful AI models. We are committed to diligently monitoring relevant research and seamlessly integrating exemplary advancements into our existing toolchain to enhance its performance.

\newpage

\bibliography{Predbenchmark} 

\end{document}